\documentclass[lettersize,journal]{IEEEtran}
\usepackage{amsmath,amsfonts}

\usepackage{algorithm}
\usepackage{array}
\usepackage[caption=false,font=footnotesize,labelfont=rm,textfont=rm]{subfig}
\usepackage{textcomp}
\usepackage{stfloats}
\usepackage{url}
\usepackage{verbatim}
\usepackage{graphicx}
\usepackage{cite}
\usepackage{threeparttable}
\usepackage{tikz}
\usepackage{colortbl,xcolor,array}
\usepackage{multicol}
\usepackage{multirow}
\usepackage{float}
\usepackage{booktabs}
\usepackage{algpseudocode}

\usepackage{algorithmicx}  
\usepackage[switch]{lineno}

\begin{document}
\title{Spiking Variational Graph Representation Inference for Video Summarization}

\author{Wenrui Li,
        Wei Han,
        Liang-Jian Deng,~\IEEEmembership{~Senior Member,~IEEE,}\\
        Ruiqin Xiong,~\IEEEmembership{~Senior Member,~IEEE,}
        Xiaopeng Fan,~\IEEEmembership{~Senior Member,~IEEE}
        
\thanks{This work was supported in part by the National Key R\&D Program of China (2023YFA1008500), the National Natural Science Foundation of China (NSFC) under grants 624B2049 and U22B2035. (Corresponding author: Xiaopeng Fan.)}
\thanks{Wenrui Li, Wei Han, and Xiaopeng Fan are with the Department of Computer Science and Technology, Harbin Institute of Technology, Harbin 150001, China, and also with Harbin Institute of Technology Suzhou Research Institute, Suzhou 215104, China. (e-mail: liwr@stu.hit.edu.cn; 2021111641@stu.hit.edu.cn; fxp@hit.edu.cn).}
\thanks{Liang-Jian Deng is with the School of mathmatical Sciences/Multi-Hazard Early Warning Key Laboratory of Sichuan Province, University of Electronic Science and Technology of China, Chengdu, Sichuan, 611731, China (e-mails: liangjian.deng@uestc.edu.cn).}
\thanks{Ruiqin Xiong is with the School of Electronic Engineering and Computer Science, Institute of Digital Media, Peking University, Beijing 100871, China (e-mail: rqxiong@pku.edu.cn)}
}



\maketitle

\begin{abstract}
With the rise of short video content, efficient video summarization techniques for extracting key information have become crucial. However, existing methods struggle to capture the global temporal dependencies and maintain the semantic coherence of video content. Additionally, these methods are also influenced by noise during multi-channel feature fusion. We propose a Spiking Variational Graph (SpiVG) Network, which enhances information density and reduces computational complexity. First, we design a keyframe extractor based on Spiking Neural Networks (SNN), leveraging the event-driven computation mechanism of SNNs to learn keyframe features autonomously. To enable fine-grained and adaptable reasoning across video frames, we introduce a Dynamic Aggregation Graph Reasoner, which decouples contextual object consistency from semantic perspective coherence. We present a Variational Inference Reconstruction Module to address uncertainty and noise arising during multi-channel feature fusion. In this module, we employ Evidence Lower Bound Optimization (ELBO) to capture the latent structure of multi-channel feature distributions, using posterior distribution regularization to reduce overfitting. Experimental results show that SpiVG surpasses existing methods across multiple datasets such as SumMe, TVSum, VideoXum, and QFVS. Our codes and pre-trained models are available at \url{https://github.com/liwrui/SpiVG}.
\end{abstract}


\section{Introduction}
\label{sec:intro}
In recent years, short video content on social media platforms has experienced rapid growth, with platforms like TikTok, Instagram Reels, and YouTube Shorts generating billions of daily video views. The global market for short video platforms is expected to reach \$3.24 billion by 2030, with a compound annual growth rate (CAGR) of 10.2\%. In this context, research into video summarization techniques has gained substantial relevance. Video summarization seeks to extract key information from lengthy video content, enabling users to grasp the main points quickly. Two primary approaches to video summarization are key-frame-based and key-shot-based methods. Key-frame-based methods create a static storyboard by selecting a representative set of frames from the video, providing an overview of the main content. In contrast, key-shot-based methods divide the video into a series of shots and select the most informative ones to produce a dynamic video skim. Dynamic video summaries are generally more engaging and enjoyable for users than static frame slideshows. Consequently, this study focuses on creating customized dynamic video skims based on user queries, aiming to capture critical information and the temporal structure of video content more accurately and comprehensively.
\begin{figure}
    \centering
	\includegraphics[width=1.0\linewidth]{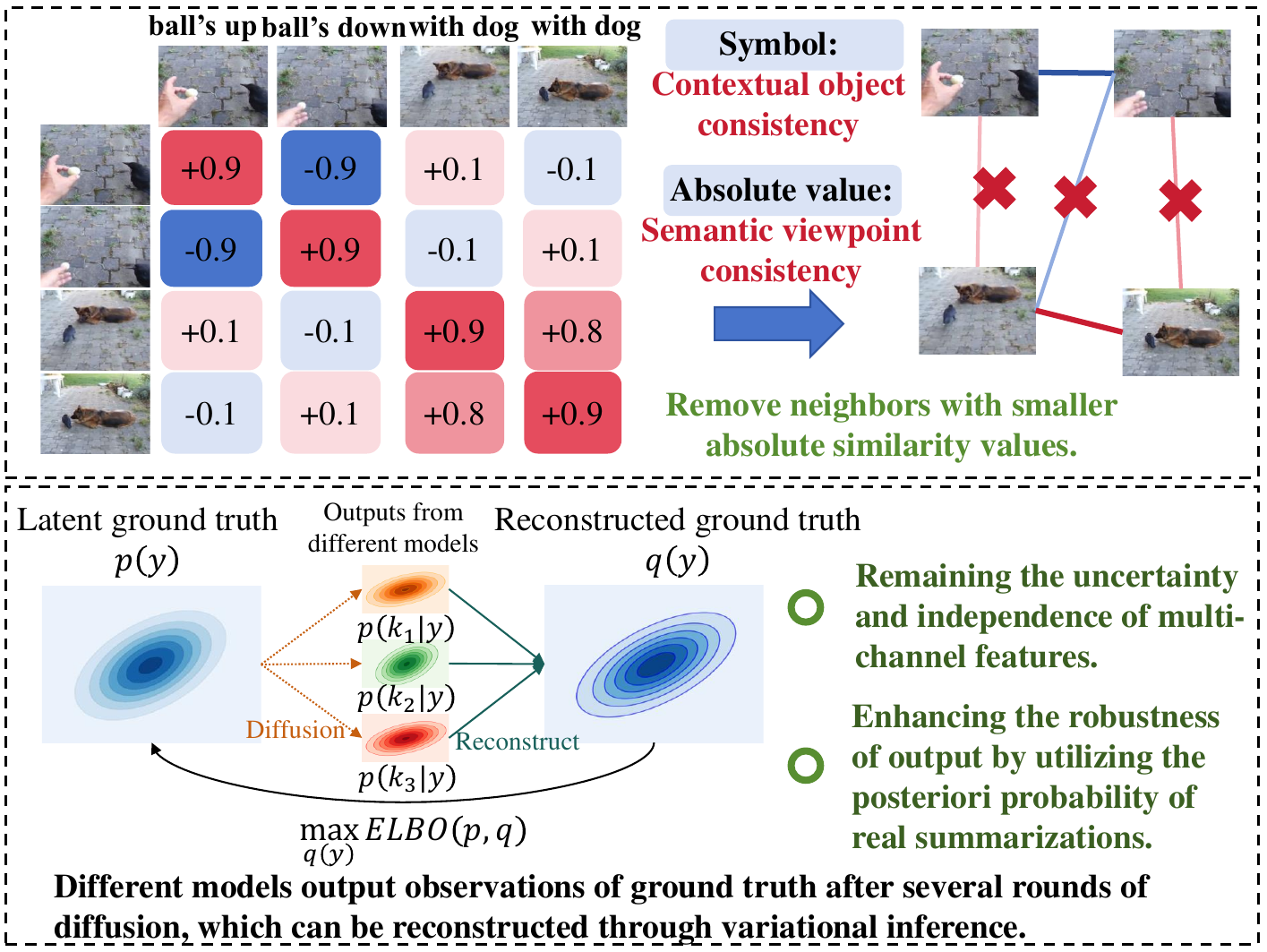}
	\caption{The absolute value of the similarity between frames indicates whether the described objects are identical. Only frames with sufficiently large absolute similarity values are retained as neighbors. Since the abstracts extracted by different models can be viewed as diffusion results of the true abstracts, variational inference is employed to reconstruct the original abstracts.}
	\label{fig:1}
\end{figure}

Current video summarization methods \cite{park2020sumgraph,desnet} primarily focus on generating multi-modal summaries that are content-rich and information-dense. He et al. \cite{intro1} propose a unified transformer-based model with contrastive loss to capture inter- and intra-sample correlations. While these methods \cite{intro2} emphasize semantic relevance, they remain limited in capturing the global, long-term dependencies essential for summarizing lengthy videos. Zhao et al. \cite{intro3} combine LSTM to capture frame-level dependencies with GCN to model shot-level interactions, enabling local and global content comprehension. Zhu et al. \cite{intro4} use intra-block and inter-block attention mechanisms to capture short-range and long-range temporal representations. Although these methods have advanced in modeling global contextual relationships, they still rely on single-modality video summaries, limiting adaptability to user-driven queries. Clip-it \cite{intro5} uses a language-guided multi-modal transformer to score frames based on importance and relevance to user-defined queries or automatically generated captions. Unlike the abovementioned methods, this study focuses on leveraging event-driven SNNs and dynamic graph reasoning. Our method decouples contextual object consistency from semantic coherence, enhancing robustness and computational efficiency while offering a more adaptive and semantically rich framework for video summarization.

As shown in Fig. 1, the absolute values of inter-frame similarity represent contextual object consistency, while their sign indicates semantic viewpoint consistency. Leveraging this property, we adjust aggregation function weights in the Graph Neural Network (GNN) to retain only neighboring nodes with strong positive or negative correlations to the central node. This filtering approach enables leveraging inter-frame similarity to differentiate between contextual object consistency and semantic viewpoint consistency, allowing further exploration of their intrinsic relationship. We use the Evidence Lower Bound (ELBO) for feature reconstruction to preserve the independence and uncertainty of multi-channel features. Assuming the ground-truth summary sequence can be regarded as a latent state $z$, we approximate its distribution by selecting the Gaussian distribution closest to the Spiking Neural Network (SNN) output. The binary output characteristic of SNNs offers a natural and effective mechanism for keyframe selection. Unlike conventional neural networks that typically rely on manually defined thresholds to decide whether a frame should be selected, SNNs inherently generate discrete spike responses driven by learned membrane potential dynamics. This enables our model to automatically learn when a frame contains sufficient information to be selected as a keyframe, eliminating the need for hand-crafted rules. Consequently, SNNs function as adaptive and interpretable selectors that enhance both efficiency and robustness, particularly in dynamic video environments. This is combined with three-way joint features from the GNN output to obtain an analytical posterior solution for $z$. By introducing ELBO, the model captures diverse information from multiple feature extraction pathways, producing independent and enriched video summaries. Each feature extraction pathway emphasizes distinct semantic or structural elements, such as scene transitions or keyframes, allowing the model to integrate meaningful features from multiple perspectives and create a comprehensive summary. ELBO optimizes posterior probability estimation of the ground-truth summary, reducing inconsistencies across multi-channel outputs, enhancing final output robustness, and improving computational efficiency.

This paper proposes a novel Spiking Variational Graph (SpiVG) network for video summarization. Specifically, to enhance the effectiveness and information density of video summaries, keyframe selection is used to reduce redundant data and lower model computational complexity. We design a spiking keyframe extractor leveraging SNN to automatically learn keyframe features without manually setting thresholds. The event-driven efficiency of SNN computation enables flexible and efficient keyframe extraction. Additionally, a Dynamic Aggregation Graph Reasoner is proposed to decouple contextual object consistency and semantic viewpoint consistency between video frames. This module separates multi-level contextual information, enabling finer-grained and more adaptive inter-frame reasoning. To efficiently integrate multi-channel features and enhance model output robustness, a Variational Inference Reconstruction Module is introduced to address potential uncertainty and noise arising during multi-channel feature diffusion. Since the covariance matrix of multi-channel features is unknown and variance increases with diffusion time, the keyframe signal may resemble noise after prolonged diffusion. Using Evidence Lower Bound Optimization (ELBO), the model captures the underlying structure of the feature distribution and regularizes the posterior distribution to mitigate overfitting. This module estimates feature sequence diffusion using the k-th order differential absolute mean across channels. It transforms these feature weights through a fully connected layer, enabling the model to adapt more effectively to varying levels of diffusion. The main contributions can be summarised as follows:
\begin{itemize}
\item This paper presents a novel Spiking Variational Graph (SpiVG) network, designed for adaptive and flexible summarization that enhances the information density of video summaries while reducing computational complexity.

\item We propose a Dynamic Aggregation Graph Reasoner that decouples contextual object consistency from semantic viewpoint consistency across video frames, enabling the model to perform adaptive inter-frame reasoning and enhancing the semantic relevance of summaries.

\item We introduce a Variational Inference Reconstruction Module that fuses summary structures from multiple feature extraction pathways. Utilizing the ELBO method, the model captures diverse information while mitigating noise and uncertainty during feature diffusion.
\end{itemize}
The extensive experimental results prove that SpiVG shows superiorities among state-of-the-art methods. The ablation study also demonstrates the effectiveness of each key component of our proposed model. The rest of the paper is organized as follows: Section \ref{sec:related} provides an overview of the background on supervised and unsupervised video summarization, along with spiking neural networks. Section \ref{sec:method} elaborates on the proposed SpiVG architectures in detail. Section \ref{sec:expe} presents the experimental results and visualizations, demonstrating the model's effectiveness. Finally, Section \ref{sec:conclu} summarizes this study's key findings and contributions.

\section{Related Work}
\label{sec:related}
\subsection{Video Summarization}
With the development of deep learning \cite{zhuangzhuang1,zhuangzhuang2,zhuangzhuang3,zhuangzhuang4,zhuoyuan2,zhuoyuan3,qianyue1,qianyue2}, event-driven representations such as spike video events \cite{zhaorui1,zhaorui2,zhaorui3,zhaorui4,zhaorui5} have also attracted increasing attention due to their efficiency and temporal precision. Building on these advances, video summarization methods can be broadly categorized into supervised and unsupervised approaches. Supervised methods \cite{s1,s2,s3,s4,s5,TIP1,TIP2,TIP3,vlm_videoxum} typically rely on manually annotated summaries as training references to build models that can generate concise, informative summaries of unseen videos. A foundational approach involves scoring each frame or shot based on its relevance and selecting the highest-scoring segments for the summary. Ji et al. \cite{supervised1} leverage an attentive encoder-decoder architecture based on BiLSTM and dual attention mechanisms to accurately model human-like keyshot selection. Zhao et al. \cite{supervised2} present a structure-adaptive video summarization method to integrate shot segmentation and summarization into a Hierarchical Structure-Adaptive RNN framework. Zhao et al. \cite{supervised3} introduce a tensor-train hierarchical recurrent neural network to address the inefficiency of large feature-to-hidden matrices in video summarization. Zhang et al. \cite{supervised4} enhance traditional discriminative losses with a metric learning loss, leveraging a “retrospective encoder” to ensure the semantic alignment between predicted summaries and original videos.

Unsupervised video summarization methods \cite{u1,u2,u3,u4} avoid the reliance on labeled data, typically transforming video summarization into a subset selection problem. Elhamifar et al. \cite{u3} introduce a sparse optimization-based framework for selecting representative data points. The framework effectively handles diverse data distributions, identifies outliers, and processes large datasets efficiently, with applications in video summarization. Lu et al. \cite{u4} propose a static video summarization method based on the Bag-of-Importance (BoI) model, which selects keyframes by evaluating the importance of local features. Jung et al. \cite{un1} combine a variance loss term to improve feature learning with a chunk-and-stride framework for long video sequences. He et al. \cite{un2} use attentive conditional GANs to incorporate frame-level multi-head self-attention to capture long-range temporal dependencies. Yuan et al. \cite{un3} leverage a cycle-consistent adversarial LSTM framework, enabling it to maximize information retention and summary compactness through a bi-directional LSTM frame selector. Fu et al. \cite{un4} utilize an attention-aware Ptr-Net generator to identify summary segments and a 3D CNN discriminator to distinguish real and generated summaries. 

Different from conventional video summarization methods, our approach leverages GNNs to ensure contextual object consistency and semantic viewpoint alignment. Furthermore, we utilize the ELBO to model the uncertainty and independence of multi-channel features, thereby improving the network's computational efficiency and robustness.

\subsection{Spiking Neural Network}
In recent years, Spiking Neural Networks (SNNs), a computational model inspired by biology, have gained significant attention. Unlike traditional Artificial Neural Networks (ANNs), SNNs use discrete spiking signals to transmit information, closely mimicking biological neural mechanisms. This approach provides potential advantages in energy efficiency and temporal coding \cite{deepsnn1,deepsnn2,sup1,sup2,sup3}. Significant progress has been achieved in theoretical frameworks and algorithm development \cite{deepsnn3,deepsnn4,spikingjelly}. For example, neuron models based on membrane potential dynamics, such as the Leaky Integrate-and-Fire (LIF) model, are widely used in SNNs to enable biologically plausible temporal sequence processing. In this model, the membrane potential $V(t)$ of a neuron changes over time according to the following differential equation:
\begin{equation} \tau \frac{dV(t)}{dt} = - \left( V(t) - V_{\text{rest}} \right) + I(t), \end{equation}
where $\tau$ represents the membrane time constant, $V_{\text{rest}}$ is the resting potential, and 
$I(t)$ is the input current. When $V(t)$ reaches a threshold $V_{th}$, the neuron emits a spike, and $V(t)$ is reset to a lower potential. By encoding spatiotemporal features as spike sequences, SNNs uniquely excel in time-dynamic modeling, making them particularly suitable for processing continuous temporal data, such as video streams. SNNs utilize their high temporal resolution via temporal coding to capture dynamic frame-to-frame changes, enabling efficient sequential modeling \cite{audiosnn1,audiosnn2,zhuSNN}. In multi-modal analysis, SNNs offer innovative solutions for synergistically processing multi-source data \cite{multisnn1,li2023modality}, leveraging their inherent adaptability to asynchronous inputs. For instance, researchers have utilized SNNs to simultaneously process visual and auditory signals, achieving cross-modal feature integration by aligning spike sequences temporally \cite{10608071,li2024spikembamultimodalspikingsaliency}.

\begin{figure*}
    \centering
	\includegraphics[width=1.0\linewidth]{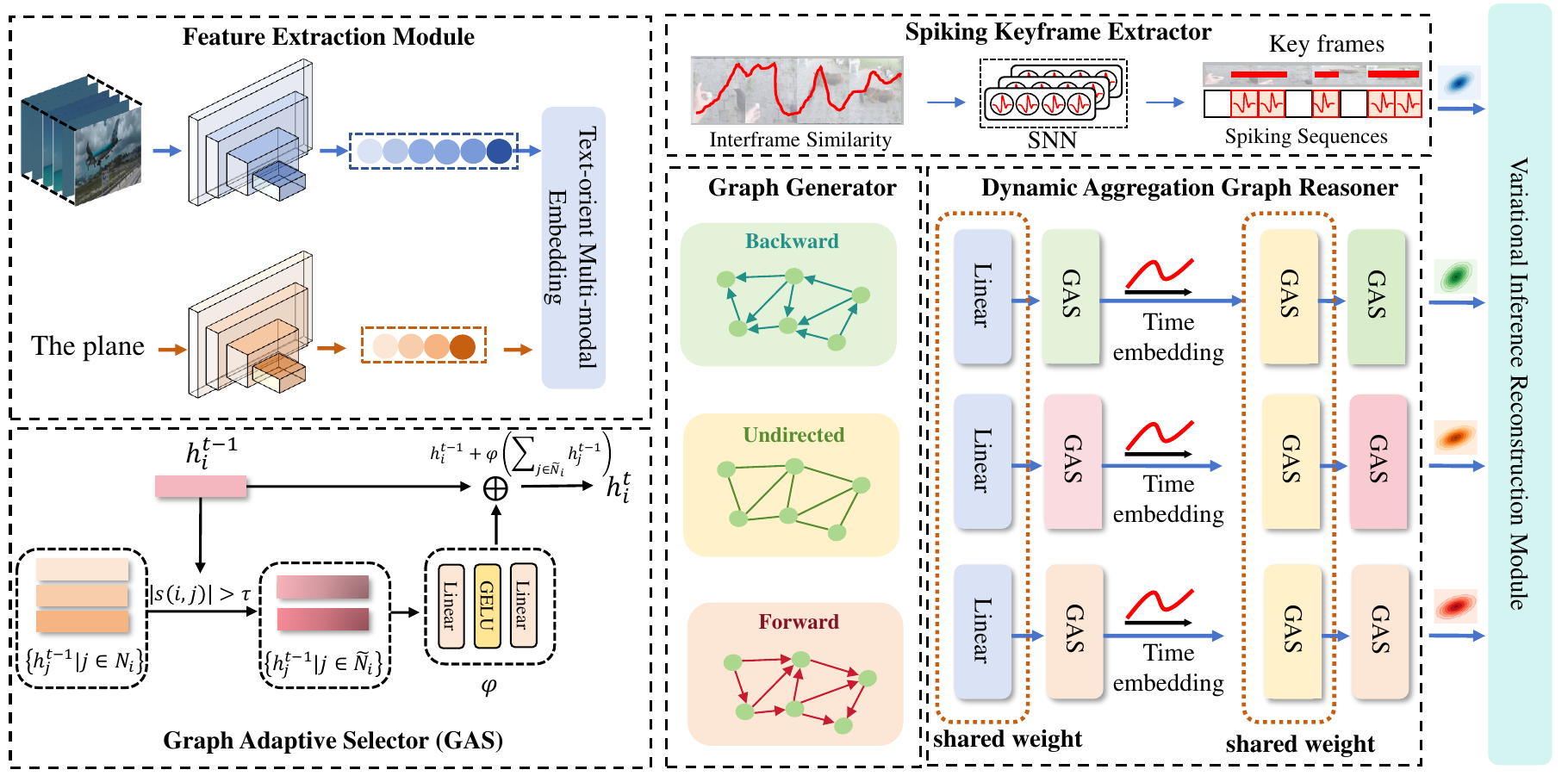}
	\caption{The overall structure of SpiVG involves extracting video features frame by frame and aggregating them with text features through the Text-Oriented Multi-Modal Embedding module. Next, the Spiking Keyframe Extractor identifies keyframes using spiking dynamics, enabling efficient and biologically inspired event-based selection. The Dynamic Aggregation Graph Reasoner, utilizing GAS, operates from three semantic perspectives. Finally, the Variational Inference Reconstruction Module synthesizes the final summary based on the results from all perspectives.}
	\label{fig:2}
\end{figure*}
\section{Methods}
\label{sec:method}

Video summarization requires identifying keyframes that capture critical content while balancing diverse factors. These factors include abrupt shot transitions and semantic cues that indicate a frame’s relevance to the video’s core message. Existing methods rely on rigid, monolithic models that cannot address the nuanced interplay among these elements. Specifically, these models struggle to adapt to varying temporal structures, integrate heterogeneous features, and account for contextual dependencies across frames. To address these challenges, we introduce the Spiking Variational Graph (SpiVG) Network. This approach leverages spiking neural dynamics for efficient keyframe extraction and uses dynamic graph reasoning to separate contextual object consistency from semantic coherence. The framework of SpiVG is shown in Fig. \ref{fig:2}.

\subsection{Task Description}

\begin{table}
\centering
		\renewcommand\arraystretch{1.4}
\caption{Key Notations and Descriptions}
\label{TAB1}
\begin{tabular}{|c|c|}
\hline
\textbf{Notation} & \textbf{Description} \\ \hline
$ V = \left\{ I_i \right\}_{i \le T} $ & Video sequence to be summarized \\ \hline
$ S \subset V $ & Summary result \\ \hline
$ x_i \in \mathbb{R}^{d_x} $ & Feature vector of the $i$-th frame \\ \hline
$ y \in \mathbb{R}^{T} $ & Summary scores \\ \hline
$ \psi^{j} $ & Map of $j$-th layer SNN network \\ \hline
$ v_t \in \mathbb{R}^{d_s} $ & Membrane potential of spiking neural network at time $t$ \\ \hline
$ N_i \subset \mathbb{Z}^+ $ & The neighbor set of the $i$-th node in GNN \\ \hline
$ h_i^t \in \mathbb{R}^{d_t} $ & The hidden vector of the $i$-th node in the t-th layer \\ \hline
$ \left \| \cdot  \right \|  $ & Euclidean norm of vectors \\ \hline
$ v^{(v)} $, $ v^{(t)} $ & Video node and text node \\ \hline
\end{tabular}
\end{table}

The video summarization task aims to identify a mapping $ f: \mathbb{R}^{\left| V \right | } \rightarrow \mathbb{R}^{\left| S \right | } $, where the input is a video sequence $V$ and the output is a subset $S$ of $V$. This subset $S$ should encapsulate the primary semantic content of $V$. Within the model-based framework, the video $V$ is initially decomposed into its constituent frames to derive a feature sequence $ \left \{ x_i \right \}_{i \in V}  $. Subsequently, a neural network is trained on the dataset to approximate the mapping $f$. In the training corpus, each video sequence is accompanied by $k$ video summary variations $ \left \{ y_i \right \}_{0 \le i < k} $, crafted by diverse annotators. Each annotation sequence $y$ comprises a binary string, where a value of 1 at the $j$-th position signifies that the $j$-th frame pertains to the summary subset $S$, and a value of 0 otherwise. In contrast to conventional video summarization tasks, video summarization guided by natural language incorporates textual information $t$ as an auxiliary input to the mapping $f$, thereby influencing its behavior. Consequently, the mapping $f$ must accommodate more intricate multi-modal data. Detailed symbol definitions are shown in Table \ref{TAB1}.

\subsection{Spiking Keyframe Extractor}
Video lens shake (such as jumps and other distinctive features) often convey semantic behaviors like changes and emphasis within video content. These behaviors are crucial in determining whether specific frames can serve as summary frames. By evaluating if fluctuations in the video feature sequence $ \left \{ x_t \right \} _{0 < t < T} $ exceed a defined threshold $ \tau_v $, one can determine if specific frames qualify as keyframes.


Euclidean distance is a fundamental metric to measure variation between adjacent frame features within the video sequence, enabling a quantitative assessment of temporal changes. The primary goal of this approach is to detect abrupt transitions, or “jumps,” between frames, which serve as strong indicators of key events or scene shifts. Notably, this method does not involve an in-depth analysis of the content differences between consecutive frames; instead, it focuses on whether a transition surpasses a predefined variation threshold. The approach maintains computational efficiency by simplifying the detection process to a numerical comparison while effectively capturing moments of significant change. 

The Euclidean distance sequence is transformed into a binary sequence to facilitate a structured representation of frame variations. Each element in this sequence denotes whether the difference between adjacent frames exceeds a predefined threshold $ \tau_v $. This binary conversion simplifies downstream processing by reducing the complexity of continuous-valued feature differences into a discrete classification problem. Additionally, achieving a specific level of temporal information fusion is crucial for ensuring that the identified keyframes mark significant transitions and maintain contextual consistency within the summarized video. Additionally, to achieve a specific level of temporal information fusion. Considering these two objectives, a spiking neural network $ \psi^{n} $ is utilized to transform the video feature difference sequence $ \left \{ \Delta x_t \right \} _{0 < t < T} $ into a keyframe sequence $ k_t $, as follows:

\begin{equation}
    \begin{aligned}
        &\Delta x_t = \left \| x_{t+1} - x_t \right \|, \\
        &k_t = \psi^{n} (t, \left \{ \Delta x_t \right \}_{0<t<T}, \tau_v ),
    \end{aligned}
\end{equation}
where $ \left \{ \Delta x_t \right \}_{0<t<T} $ is the video feature differential sequence, $ t $ is the current time, $ k_t \in \left \{ 0, 1 \right \} $ is whether time $t$ is a keyframe or not. The spiking neural network $ \psi^{n} $ uses a $n$ layers LIF model, as follows:
\begin{equation}
    \begin{aligned}
        &\psi^{0}(t, \left \{ \Delta x_t \right \}_{0<t<T}, \tau_v) = \Delta x_t, \\
        &\psi^{n}(t,\cdot, \tau_v)  = \psi(t, \psi^{n-1}(t, \cdot, \tau_v), \tau_v), \\
        &\psi(t, \left \{ z_t \right \}_{0<t<T}, \tau_v)=\left\{\begin{matrix} 0, v_t<\tau_v
 \\1, v_t\ge \tau_v
\end{matrix}\right.\\
        &C_m (v_t-v_{t-1}) = -G_L(v_{t-1}-E_L) + z_t,
    \end{aligned}
\end{equation} 
where $C_m$ denotes the membrane potential, and $G_L$ and $E_L$ denote admittance and resting potential, respectively. $ z_t $ represents the input spiking signal, and $ v_t $ represents the output voltage. Additionally, the voltage $v_t$ should stay at the reset potential for a specified period after emitting a spike. 

The introduction of spiking neural networks in keyframe selection provides a biologically inspired and computationally efficient mechanism for processing video sequences. However, isolated keyframe detection does not fully address the complexities of video summarization, as maintaining contextual coherence and capturing long-range dependencies are equally critical. A Dynamic Aggregation Graph Reasoner is introduced in the following subsection to further refine the summarization process. This module extends the SNN-based keyframe extraction by leveraging graph-based reasoning, allowing the model to distinguish between contextual object consistency and semantic coherence across frames. The summarization framework achieves a more comprehensive and context-aware representation of video content by integrating spiking keyframe extraction with graph-based semantic analysis.

\subsection{Dynamic Aggregation Graph Reasoner}
To capture semantic features from the video sequence $ \left \{ x_t \right \} _{0 < t < T} $, we model the sequence as three distinct graphs: a forward $ G^f $, a backward $ G^b $, and an undirected $ G^u $. Leveraging graph neural networks, the semantic information is extracted from these graphs to produce a comprehensive summary. In constructing the graph, an edge between any two frames is established when their time interval is below a specified threshold. For the undirected graph $G^u$, these edges lack direction and are considered dimensionless. In contrast, in the forward graph $G^f$, edges are directed from earlier to later frames, while in the backward graph $G^b$, edges point from later to earlier frames. The reason for modeling video sequences as three types of graphs is that undirected graphs can bring information about the preceding and following frames for each frame, but ignore their order; Forward graph can maintain the order of video playback, but cannot bring information after each frame; The backward graph can bring subsequent information, but it loses the preceding information.


A key challenge in video summarization is managing the high redundancy in video frames, particularly due to the high sampling rates of modern video datasets. Adjacent frames often exhibit substantial similarity, resulting in high neighborhood similarity within constructed graphs. Traditional graph neural networks, such as GCN and GraphSAGE, typically aggregate information by weighing and summing the features of neighboring nodes. While effective in capturing local relationships, this approach inadvertently leads to repeated extraction of redundant information. Consequently, to mitigate redundant aggregation, we propose a redesigned graph neural network architecture incorporating an adaptive neighbor filtering mechanism. Specifically, we assign an aggregation weight of zero to uninformative neighbors, effectively pruning redundant connections and ensuring that only meaningful contextual relationships are retained during feature aggregation. This adjustment enhances the model’s efficiency by reducing computational overhead while preserving key structural patterns within the video.

Specifically, we first apply a fully connected layer $f$ to extract semantic information $f(h_i^{t-1})$ related to a particular aspect of node $h_i^{t-1}$ within the $(t-1)$-th layer of the graph. Subsequently, the cosine similarity between each node $i$ and all its neighboring nodes $ N_i $ is computed. Nodes with absolute similarity values below a certain threshold are excluded, and the mean of the remaining neighbors $ \widetilde{N}_i^{t} $ serves as the aggregation result. A similarity of 0 indicates that these neighbors and node $h$ are semantically orthogonal in this aspect, suggesting they do not describe the same entity. The symbol of similarity represents whether their descriptions of the same semantic object are consistent. We can use the mean to summarize the viewpoints of neighboring nodes and pass them on to the target node. The formal description of this module is as follows:
\begin{equation}
    \begin{aligned}
        &\widetilde{N}_i^{t} = \left \{ j|j \in N_i \wedge \frac{\left | f(h_i^{t-1})'f(h_j^{t-1}) \right | }{\left \| f(h_i^{t-1}) \right \| \left \| f(h_j^{t-1}) \right \|  }>\tau_v  \right \}, \\
        &h_i^t=\varphi (\frac{1}{\left | \widetilde{N}_i^{t} \right | } {\textstyle \sum_{j \in \widetilde{N}_i^{t}} h_j^{t-1}}) + h_i^{t-1},
    \end{aligned}
\end{equation}
where $\varphi(\cdot)$ is a small feedforward network composed of two fully connected layers and a GELU activation function. While the Dynamic Aggregation Graph Reasoner enhances the extraction of meaningful relationships between frames, it does not account for the inherent uncertainty and noise introduced during multi-channel feature fusion. The variability in extracted features across different summarization pathways necessitates a robust mechanism for regularizing and reconstructing the underlying summary representation. To address this challenge, we employ Evidence Lower Bound Optimization (ELBO) to approximate the latent structure of multi-channel feature distributions, mitigating overfitting and enhancing the robustness of the summarization output. By integrating probabilistic inference with graph-based reasoning, the model achieves a more adaptive and resilient representation of video content, ensuring both semantic coherence and structural consistency in the generated summaries.
\subsection{Variational Inference Reconstruction Module}


Variational inference is employed to derive the final summary result from summary structures obtained from various sources. The summary result from the $i$-th route is defined as $k_i$, while the actual summary result is represented by $y$. Since each path may emphasize different abstract aspects, we assume that $k_i$ results from the diffusion of $y$ over a specific time period. Consequently, $k$ follows a Gaussian distribution, with $y$ serving as its expected value:
\begin{equation}
    p(k_i|y)=N(k_i|y,\Sigma_i).
\end{equation}

Assuming no overlap between the focuses, the summary results for the $N$ focuses, $k_{0}$, are independent of each other:
\begin{equation}
    p(k_{0:N}|y) = {\textstyle \prod_{i} p(k_i|y)}.
\end{equation}

Therefore, given a Gaussian prior distribution with mean $ \mu_0 $ and covariance matrix $ \Sigma_0 $ for $y$, and considering $y$ as a hidden state and $k_{0}$ as observable states, the variational inference is used to determine the posterior probability of $y$. In variational inference, we aim to maximize the Evidence Lower Bound (ELBO) by adjusting $q(y)$:
\begin{equation}
    ELOB= \int q(y)\mathrm{log}(\frac{p(k_{0:N}, y)}{q(y)})dy.
\end{equation}

By substituting the previous assumption of independence, the following equation can be defined:
\begin{equation}
    \begin{aligned}
        &\int q(y)\mathrm{log}(\frac{p(k_{0:N}, y)}{q(y)})dy = \\
        &\mu_\theta'(\Sigma_0^{-1}+ {\textstyle \sum_{i}\Sigma_i^{-1}+\Sigma_y^{-1}})\mu_\theta \\
        &-2(\mu_0'\Sigma_0^{-1}+ {\textstyle \sum_{i}k_i'\Sigma_i^{-1}})\mu_\theta +C.
    \end{aligned}
\end{equation}

Find a stationary point and obtain:
\begin{equation}
    \begin{aligned}
        \mu_{\theta}=
        (\Sigma_0^{-1}+ {\textstyle \sum_{i}\Sigma_i^{-1}+\Sigma_y^{-1}})^{-1} (\Sigma_0^{-1}\mu_0+ {\textstyle \sum_{i}\Sigma_i^{-1}k_i}),
    \end{aligned}
\end{equation}
where each $ \Sigma_i $ is a diagonal matrix, with its diagonal element $\sigma_i$ indicating the diffusion degree of the path. Assume that this degree can be approximated from the inherent properties of $k_i$, as an elongated diffusion sequence increasingly resembles a stationary sequence. The average absolute value of the $n$ differences in the $k_i$ sequence can be used as a feature to estimate $\sigma_i$, specifically as follows:
\begin{equation}
    \begin{aligned}
        \delta_i^m= \frac{1}{N} {\textstyle \sum_{t}\left | (k_i)_{t+m} - (k_i)_{t} \right | } \\
        \sigma_i^2=e^{ b_i+{\textstyle \sum_{m}w_m\delta_i^m} },
    \end{aligned}
\end{equation}
where $\delta_i^m$ denotes the $m$-order absolute mean of the $i$-th channel difference, $(k_i)_{t}$ represents the $t$-th element of the $k_i$ sequence, and $b_i$ is a bias term representing the intrinsic components of the $i$-th channel.

\subsection{Text-orient Multi-modal Embedding}
SpiVG incorporates a text-oriented multi-modal embedding module to improve its suitability for natural language-guided video summarization. This enhancement allows it to produce summaries that more accurately align with user-defined textual queries $T$. Since SpiVG is inherently designed for video sequence processing, an effective mechanism is needed to integrate textual information into the summarization pipeline while ensuring semantic alignment between text $ T $ and visual content $ V $. To achieve this, we pre-integrate textual features into the video sequence and seamlessly incorporate them into SpiVG’s architecture. This is achieved through a heterogeneous graph representation that structures textual and visual data within a unified framework, facilitating cross-modal feature interactions. Specifically, textual information is first encoded with BERT, generating a dense and semantically rich representation of the input query. This encoded text information is then incorporated as a dedicated text node \( v^{(t)} \) within the heterogeneous graph. Simultaneously, each individual video frame is represented as a video node \( v^{(v)} \), preserving the temporal structure of the video sequence. The above heterogeneous graph $ G(V, E) $ can be formally described as:
\begin{equation}
    \begin{aligned}
        V = \left \{ v^{(v)}_i \right \}_{i \in V} \cup \left \{ v^{(t)}_i \right \}_{i \in T} \\
        E = \left \{ (i, j) | v_i \in v^{(t)} \wedge v_j \in v^{(v)} \right \} 
    \end{aligned}
\end{equation}

Unlike traditional multi-modal approaches that rely on direct feature concatenation, our method explicitly models the relationships between video and text by establishing structured graph-based connections. This can artificially control the direction of information flow during information fusion, thereby enhancing the interpretability of model behavior by controlling special fusion behaviors.

We employ a key design principle during the text-information fusion process: internal associations within the same modality are disregarded. This means that no direct edges are created between video nodes, preventing excessive redundancy in modeling intra-video dependencies already handled by the Dynamic Aggregation Graph Reasoner. Instead, edges are selectively established between the text node \( v^{(t)} \) and every video node \( v^{(v)} \), ensuring that the textual input explicitly influences each frame. This formulation allows the model to propagate semantic cues from text to visual content, effectively guiding the summarization process to highlight text-relevant keyframes while suppressing unrelated content. By leveraging this structured heterogeneous graph embedding, SpiVG enhances its ability to perform query-aware video summarization, ensuring that the generated summaries capture key events that are most relevant to the user’s intent. The incorporation of text-video connectivity strengthens cross-modal alignment, enabling SpiVG to produce more contextually rich and interpretable summaries that cater to a diverse range of natural language queries. The graph neural networks are used for feature fusion as follows:
\begin{equation}
    \begin{aligned}
        x^{(v)}=v^{(v)}+ \sigma((W_1v^{(t)})'(W_2v^{(v)}))\cdot(W_3v^{(t)}+b),
    \end{aligned}
\end{equation}
where $ x^{(v)} $ is the video node that aggregates text information $ v^{(t)} $, and $ \sigma(\cdot) $ is the Sigmoid function.

\section{Experiments}
\label{sec:expe}

\subsection{Datasets} The SumMe \cite{gygli2014creating} and TVSum \cite{song2015tvsum} datasets are used to evaluate the efficacy of our model. The SumMe dataset is a benchmark collection designed for video summarization research, comprising 25 videos with diverse content, including daily activities, sports, and travel. Each video, ranging in length from 1 to 6 minutes, is annotated with multiple human-generated summaries that reflect subjective preferences regarding important content. The TVSum dataset is a benchmark resource for video summarization research, consisting of 50 YouTube videos from 10 diverse categories, such as news, sports, and tutorials, with durations ranging from 2 to 10 minutes. Each video is annotated with frame-level importance scores derived from crowdsourcing, reflecting human judgments about key content. These annotations serve as ground truth for evaluating video summarization methods, particularly in tasks like keyframe selection, temporal segmentation, and personalized summarization.

Furthermore, we incorporated the extensive VideoXum \cite{intro9} dataset boasts over 10,000 videos, and the multi-modal video summarization dataset QFVS \cite{akhare2022query}, which is guided by natural language. The VideoXum dataset is a large-scale resource designed for cross-modal video summarization, aiming to generate both concise video clips and corresponding textual narratives from longer videos. The VideoXum dataset comprises over 14,000 long-duration, open-domain videos, each paired with ten human-annotated summaries, totaling approximately 140,000 video-text summary pairs. The QFVS dataset is a specialized benchmark designed for personalized video summarization, featuring approximately 300 hours of consumer-grade videos. Each video is segmented into 5-second shots and annotated with 48 detailed semantic concepts, covering objects, actions, and scenes. Additionally, the dataset includes query-focused summaries tailored to 46 unique text queries per video, enabling the development and evaluation of algorithms that adapt to individual user preferences.
\begin{algorithm}[t]
\caption{Training framework for the proposed approach}
\label{algo1}
\begin{algorithmic}[1]

\Require Dataset $X$, where in any item $(x, t, y)$ of the dataset, $x$ is a video, $t$ is the natural language guidance of the video summary, and $y$ is the ground truth of the manually provided video summary.

\Ensure Update network parameters.

\State Extract features from the video and text $(v, t)$ in the dataset as $(x_t, x_v)$.
\For{each epoch}
    \For{$(x_t, x_v, y)$ in $X$}
        \State $ x \gets TME(x_t, x_v) $ \Comment{Text-orient Multi-modal Embedding}
        \State $ z_0 \gets SKE(x) $ \Comment{Spiking Keyframe Extractor}
        \State $ G \gets GG(x) $ \Comment{Graph Generator}
        \State $ i \gets 1 $
        \For{$ g $ in $ G $}
            \State $ z_{i} = DAGR_{i-1}(g) $  \Comment{The i-th channel of Dynamic Aggregation Graph Reasoner}
            \State $ i += 1 $
        \EndFor
        \State $ \hat{y} \gets VIRM(z) $ \Comment{Variational Inference Reconstruction Module}
        \State $ L \gets loss(\hat{y}, y) $
        \State Optimize model parameters using optimizers such as AdamW.
    \EndFor
\EndFor
\end{algorithmic}
\end{algorithm}

\begin{table*}
	\centering
	\begin{threeparttable}
		\renewcommand\arraystretch{1.2}
		\caption{The Performance Comparison Between the SpiVG with Other Baselines in SumMe and TVSum Datasets.}
		\label{TAB2}
		\setlength{\tabcolsep}{5pt}{
\begin{tabular}{cccccclcccclcccc}
\hline \hline
  \multirow{2}{*}{Type} &
  \multirow{2}{*}{Model} &
  \multicolumn{4}{c}{SumMe} &
   & 
  \multicolumn{4}{c}{TVSum} &
   \\ \cline{3-6} \cline{8-11} 
 &
   &
  F1 score $\uparrow$ &
  Kendall’s $\tau$ $\uparrow$&
  Spearman’s $\rho$ $\uparrow$&
   &
   &
  F1 score $\uparrow$&
  Kendall’s $\tau$ $\uparrow$&
  Spearman’s $\rho$ $\uparrow$&

   \\ \hline
\multirow{1}{*}{Human\cite{intro8}} &
 - &
  54.0 &
  0.21 &
  0.21 &
   &
   &
  54.0 &
  0.18 &
  0.20 &
   &
\\ \hline
\multirow{7}{*}{Others} &
 SUM-DeepLab \cite{rochan2018video} &
  48.8 &
  - &
  - &
   &
   &
  58.4 &
  - &
  - &
   & \\
 &
 dqqLSTM \cite{zhang2016video} &
  38.6 &
  0.04 &
  0.06 &
   &
   &
  54.7 &
  - &
  - &
   & \\
 &
 DR-DSN \cite{zhou2018deep} &
  42.5 &
  0.05 &
  0.05 &
   &
   &
  58.1 &
  0.02 &
  0.03 &
   & \\
 &
  DHAVS \cite{dhvas}  &
  45.6 &
  - &
  - &
   &
   &
  \textbf{60.8} &
  0.08 &
  0.09 &
   &  \\
 &
 HSA-RNN \cite{zhao2018hsa} &
  42.3 &
  0.06 &
  0.07 &
   &
   &
  58.7 &
  0.08 &
  0.09 &
   & \\
   \hline
\multirow{3}{*}{LLM} &
 M3SUM(SP) \cite{vlm_m3sum} &
  43.6 &
  - &
  - &
   &
   &
  56.9 &
  - &
  - &
   & \\
 &
 M3SUM(CoT) \cite{vlm_m3sum} &
  41.9 &
  - &
  - &
   &
   &
  57.6 &
  - &
  - &
   & \\
\hline
\multirow{4}{*}{GNN} &
  VideoSAGE \cite{chaves2024videosage} &
  46.0 &
  0.12 &
  0.16 &
   &
   &
  58.2 &
  0.30 &
  0.42 &
   & \\
 &
  RSGN \cite{intro8}  &
  45.0 &
  0.08 &
  0.09 &
   &
   &
  60.1 &
  0.08 &
  0.09 &
   &  \\
 &

  SpiVG (ours) &
  \textbf{53.7} &
\textbf{0.15} &
  \textbf{0.20} &
   &
   &
  57.6 & 
  \textbf{0.30} &
  \textbf{0.43} &
   &  \\
  \hline \hline
\end{tabular}
}
\end{threeparttable}
\end{table*}
\subsection{Evaluation Protocol}
To ensure a comprehensive and fair assessment, we employ diverse evaluation metrics specifically tailored to the characteristics of the datasets, while maintaining consistency with previous studies. We adopt three widely recognized evaluation metrics for the SumMe, TVSum, and VideoXum datasets: The F1 Score represents the balance between precision and Recall by providing their harmonic mean, thereby assessing the overall accuracy of the generated video summaries. Kendall's $\tau$ Coefficient measures the rank correlation between the predicted importance scores and the ground truth, providing insights into how effectively the relative order of video segments is preserved. Spearman's $\rho$ Coefficient, similar to Kendall's $\tau$, also measures rank correlation but is more sensitive to larger rank differences. It complements $\tau$ by offering an alternative perspective on correlation.

For the QFVS dataset, the evaluation metrics are defined as follows: Precision refers to the proportion of correctly selected keyframes or segments among all predicted selections, whereas Recall indicates the proportion of relevant keyframes or segments that are correctly identified from the ground truth. To maintain robustness in our experimental setup, we adopt cross-validation. Specifically, the SumMe, TVSum, and VideoXum datasets are partitioned into five folds, ensuring that each fold serves as a test set once while the others are used for training and validation. For the QFVS dataset, a similar approach is followed but with four folds. This stratified division ensures that the evaluation is not biased by the specific choice of training or testing samples, enhancing the reliability and generalizability of the results. By employing these metrics and cross-validation schemes, we provide a thorough and systematic evaluation of our method across diverse datasets.

\begin{table}
	\centering
	\begin{threeparttable}
		\renewcommand\arraystretch{1.2}
		\caption{The Performance Comparison in VideoXum Dataset.}
		\label{TAB3}
		\setlength{\tabcolsep}{2pt}{
\begin{tabular}{cccccclcccclcccc}
\hline \hline
  &
  \multirow{2}{*}{Model} &
  \multicolumn{4}{c}{VideoXum} &
   \\ \cline{3-6} 
 &
   &
  F1 score $\uparrow$&
  Kendall’s $\tau$ $\uparrow$&
  Spearman’s $\rho$ $\uparrow$&
   &
   
   \\ \hline
    &
 Human \cite{intro9} &
  33.8 &
  0.305 &
  0.336 &
   &
   \\ \hline
 &
 Frozen-BLIP \cite{intro9} &
  16.1 &
  0.008 &
  0.011 &
   &
   \\
 &
 VSUM-BLIP \cite{intro9} &
  23.1 &
  0.185 &
  \textbf{0.246} &
   &
    \\
 &
  VTSUM-BLIP \cite{intro9} &
  22.7 &
  0.176 &
  0.232 &
   &
    \\
 &
  
  SpiVG (ours) &
  \textbf{31.0 }&
  \textbf{0.186} &
  0.245 &
   &
  \\
  \hline \hline
\end{tabular}
}
\end{threeparttable}
\end{table}
\begin{table*}
	\centering
	\begin{threeparttable}
		\renewcommand\arraystretch{1.4}
		\caption{The Performance Comparison Between the SpiVG with Other Baselines in QFVS Dataset.}
		\label{TAB4}
		\setlength{\tabcolsep}{4pt}{
\begin{tabular}{ccccclccclccclccclccclccc}
\hline \hline
  &
  \multirow{2}{*}{Model} &
  \multicolumn{4}{c}{Vid 1} &
  \multicolumn{4}{c}{Vid 2} &
  \multicolumn{4}{c}{Vid 3} &
  \multicolumn{4}{c}{Vid 4} &
  \multicolumn{3}{c}{Average} &
   \\ \cline{3-5} \cline{7-9} \cline{11-13} \cline{15-17} \cline{19-21} 
 &
   &
  P $\uparrow$ &
  R $\uparrow$ &
  F1 $\uparrow$&
  &
  P $\uparrow$&
  R $\uparrow$&
  F1 $\uparrow$&
  &
  P $\uparrow$&
  R $\uparrow$&
  F1 $\uparrow$&
  &
  P $\uparrow$&
  R $\uparrow$&
  F1 $\uparrow$&
  &
  P $\uparrow$&
  R $\uparrow$&
  F1 $\uparrow$&
   \\ \hline
    & SH-DPP \cite{sharghi2016query}
    & 50.56 
    & 29.64 
    & 35.67 &
    & 42.13 
    & 46.81 
    & 42.72 &
    & 51.92 
    & 29.24 
    & 36.51 &
    & 11.51 
    & 62.88 
    & 18.62 &
    & 39.03 
    & 42.14
    & 33.38 & \\
    & QC-DPP \cite{sharghi2017query}
    & 49.86 
    & \textbf{53.38} 
    & 48.68 &
    & 33.71 
    & \textbf{62.09} 
    & 41.66 &
    & 55.16  
    & 62.40 
    & 56.47 &
    & 21.39  
    & \textbf{63.12}  
    & 29.96 &
    & 40.03 
    & \textbf{60.25} 
    & 44.19 & \\
    & TPAN \cite{zhang2018query}
    & 49.66 
    & 50.91  
    & 48.74 &
    & 43.02 
    & 48.73   
    & 45.30 &
    & 58.73   
    & 56.49   
    & 56.51 &
    & \textbf{36.70 }  
    & 35.96   
    & 33.64 &
    & 47.03  
    & 48.02  
    & 46.05 & \\
    & CHAN  \cite{xiao2020convolutional}
    & 54.73 
    & 46.57   
    & 49.14 &
    & 45.92  
    & 50.26    
    & 46.53 &
    & 59.75   
    & \textbf{64.53}   
    & 58.65 &
    & 25.23   
    & 51.16    
    & 33.42 &
    & 46.40  
    & 53.13
    & 46.94 & \\
    & FCSNA  \cite{patel2024your}
    & 55.89 
    & 39.41   
    & 45.15 &
    & 47.47  
    & 54.38    
    & 50.32 &
    & 67.88   
    & 49.93   
    & 57.24 &
    & 28.22   
    & 56.40    
    & 37.20 &
    & 49.86  
    & 50.03
    & 47.47 & \\ \hline
    & SpiVG (ours)
    & \textbf{60.26}
    & 43.32
    & \textbf{49.29} &
    & \textbf{49.59} 
    &  56.64
    & \textbf{52.48} &
    &\textbf{70.85}  
    & 51.93
    & \textbf{59.62} &
    & 29.91   
    & 59.66    
    & \textbf{39.41} &
    & \textbf{52.65}  
    &  52.89
    & \textbf{50.20} & 
  \\
  \hline \hline
\end{tabular}
}
\end{threeparttable}
\end{table*}
\subsection{Implementation Details}
The complete process of the method described in this article can be expressed as Algorithm \ref{algo1}. We utilize GoogLeNet \cite{szegedy2015going} to extract video frame features from the SumMe and TVSum datasets, while employing Blip \cite{li2022blip} for feature extraction from the QFVS dataset. AdamW \cite{loshchilov2017decoupled} served as our optimizer, with a dropout rate of 0.4 and an L2 regularization term set to 0.01. For video segmentation, KTS \cite{potapov2014category} and subsequently applied the Knapsack algorithm \cite{pisinger1999core} are leveraged to select the sequence that would ultimately be deemed as the keyframe for the model's output. All learning rates in the experiment are 0.001, and AdamW's beta1 and beta2 are 0.9 and 0.999, respectively. The experiment is conducted on one NVIDIA RTX3090, and training on each dataset takes no more than 12 hours. The batch size is set to 64, running a total of 50 epochs on the TVSum dataset, 40 on SumMe, 10 on VideoXum, and 20 on QFVS.

\subsection{Performance Comparison}
\subsubsection{Performance on SumMe and TVSum datasets} We conducted comprehensive performance comparisons on the SumMe and TVSum datasets in Table \ref{TAB2}. In the SumMe dataset, our proposed SpiVG model achieves an F1 score of 53.7 and outperforms other models like SumGraph of 51.4 and RSGN of 45.0. SpiVG also surpasses competing models in Kendall’s $\tau$ and Spearman’s $\rho$, scoring 0.15 and 0.20, compared to DR-DSN’s 0.05 and 0.05, and HSA-RNN’s 0.06 and 0.07. The Variational Inference Reconstruction module effectively integrates information from multiple channels, enhancing the model’s accuracy and consistency. SpiVG also shows significant strengths in 0.3 of Kendall’s $\tau$ and 0.43 of Pearman $\rho$, outperforming SumGraph and indicating a superior alignment with human preferences in video summarization.

\subsubsection{Performance on Large Scale VideoXum Dataset}
 We conducted comprehensive performance comparisons on the VideoXum dataset in Table \ref{TAB3}. In the VideoXum dataset, the SpiVG model achieves an F1 score of 31.0. Though slightly below the human benchmark of 33.8, this score significantly outperforms models like VTSUM-BLIP of 22.7 and Frozen-BLIP of 16.1. This higher F1 score shows that SpiVG generates more accurate summaries that closely align with human-selected keyframes. Furthermore, SpiVG demonstrates strong performance in Kendall’s $\tau$ of 0.186 and Spearman’s $\rho$ of 0.245, reflecting its rank correlation with human annotations. Although slightly below the human benchmark (0.305 for Kendall’s $\tau$ and 0.336 for Spearman’s $\rho$), SpiVG outperforms other automated models. For instance, the Frozen-BLIP model scores only 0.008 and 0.011, while the next-best model, VTSUM-BLIP, achieves 0.176 and 0.232.

\subsubsection{Performance on Multi-modal Dataset}
SpiVG consistently outperforms the baselines across most metrics on the QFVS dataset in Table \ref{TAB4}. Specifically, SpiVG achieves the highest F1-scores on three of the four video sets (Vid 1, Vid 3, and Vid 4), with notable improvements in Vid 3 F1-score of 59.62 and Vid 1 F1-score of 49.29, surpassing the best baseline (CHAN) by 1.35 and 2.18 percentage points, respectively. Additionally, SpiVG achieves the highest precision (P) on Vid 3 of 70.85 and Vid 1 of 60.26, demonstrating its superior ability to select relevant frames accurately. On average, SpiVG achieves the highest F1-score of 50.20, outperforming the second-best baseline CHAN by 3.26\%.

To further assess computational efficiency, we also report the Multiply-Accumulate Operations (MACs) of our model. Since the QFVS dataset provides only pre-extracted features without raw pixel inputs, we cannot compute kMACs/pixel. Instead, we use overall MACs as a comparative metric. The proposed SpiVG model requires only 191.9 million MACs, which is significantly lower than that of the CHAN model (3755.6 million MACs). Despite this substantial reduction in computation, SpiVG outperforms CHAN across all video categories in the QFVS dataset, demonstrating that our design achieves superior summarization performance at a fraction of the computational cost.

\subsection{Convergence and Generalization Analysis}
\begin{figure}
	\centering
	\includegraphics[width=1\linewidth]{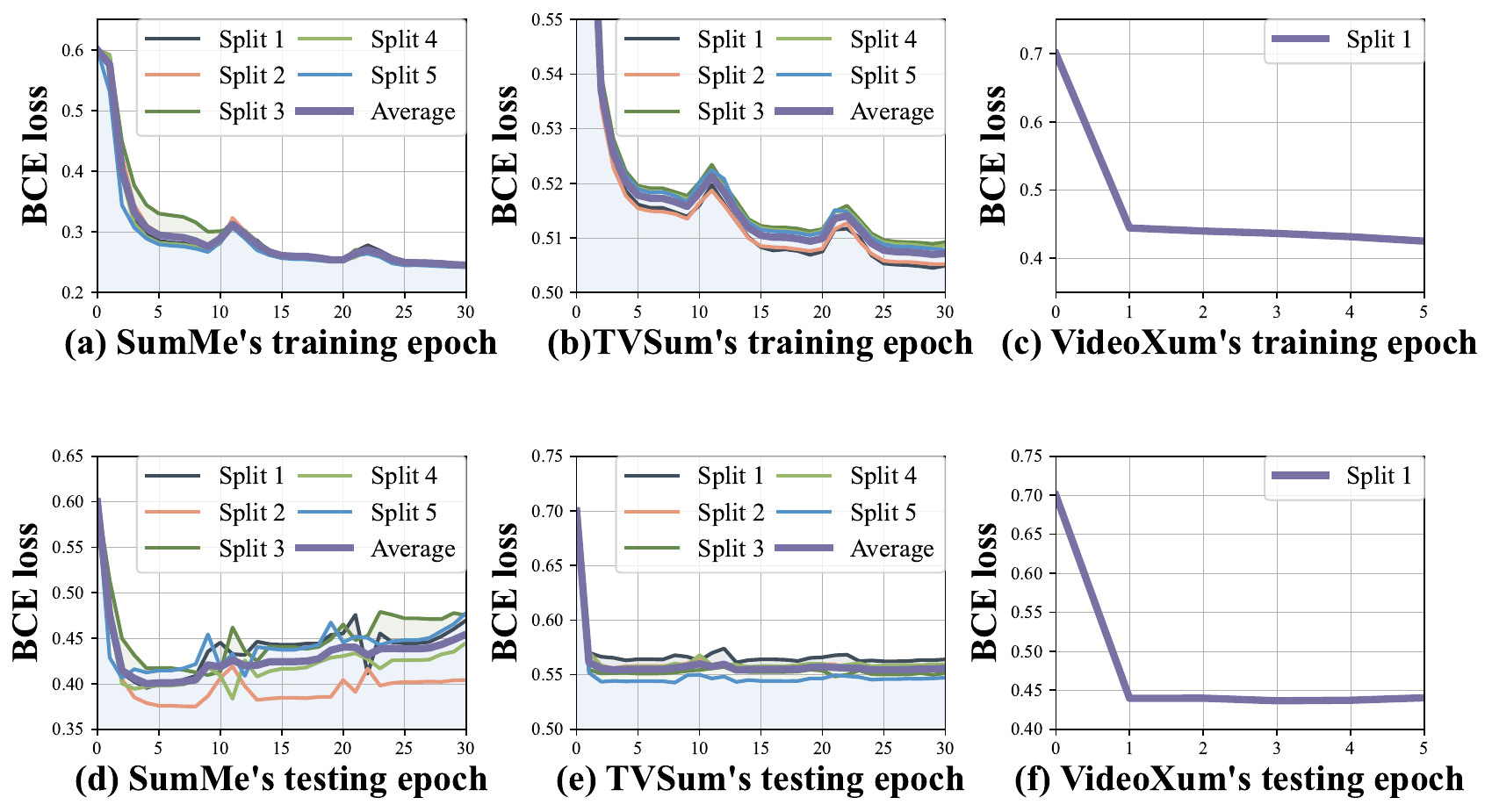}
	\caption{The convergence rates on the SumMe, TVSum, and VideoXum datasets demonstrate that our method achieves rapid convergence, reaching the optimal solution within 20 epochs.}
	\label{results}
\end{figure}
Fig. 3 illustrates the convergence and generalization capabilities of our method on three benchmark datasets: SumMe, TVSum, and VideoXum. During the training phase (Fig. 3(a)–(c)), the BCE loss decreases rapidly within the first few epochs across all datasets, highlighting the efficiency of our optimization strategy. Notably, the VideoXum dataset achieves near-optimal convergence within just five epochs, as shown in Fig. 3(c) and 3(f). For the SumMe and TVSum datasets (Fig. 3(a) and 3(b)), convergence is similarly achieved within 20 epochs, showing minimal fluctuations across different data splits, as evidenced by the average loss curves. The testing phase (Fig. 3(d)–(f)) further confirms the robustness and generalization of our method. Across all datasets, the testing loss aligns closely with the training loss, indicating minimal overfitting. Importantly, the consistent performance across multiple splits, particularly in Fig. 3(d) and 3(e), highlights the model's stability when applied to unseen data.
\begin{table}
	\centering
	\begin{threeparttable}
		\renewcommand\arraystretch{1.2}
		\caption{Ablation Study on Different SNN layer numbers.}
		\label{TAB5}
		\setlength{\tabcolsep}{5pt}{
\begin{tabular}{cccccccccccccccc}
\hline \hline
  \multirow{2}{*}{layers} &
  \multicolumn{3}{c}{SumMe} &&
  \multicolumn{3}{c}{TVSum} &
   \\ \cline{3-4} \cline{7-8} 
 &
   &
  F1-max $\uparrow$&
  F1-mean $\uparrow$&
   &
   &
  F1-max $\uparrow$&
  F1-mean $\uparrow$&
   \\ \hline
   0 &&
   44.8 &
   16.9 &
   &
   &
   80.3 &
   55.2 &\\
   1 &&
  45.3 &
  17.9 &
   &
   &
  77.9 &
  55.1 &\\
     2 &&
  59.9 &
  17.8 &
   &
   &
   78.6 &
   57.4 &\\ 
  3 &&
  50.2 &
  19.9 &
   &
   &
  78.2 &
  54.5 &\\
  \hline \hline
\end{tabular}
}
\end{threeparttable}
\end{table}

\subsection{Ablation Study}
\subsubsection{The Effectiveness of Different SNN Layer Numbers} We evaluate the effectiveness of different SNN layer numbers in Table \ref{TAB5}. On the SumMe dataset, the F1-max metric exhibits a substantial improvement as the number of SNN layers increases from 0 to 2, rising from 44.8 to 59.9. This trend indicates that integrating SNN layers enhances the model's ability to identify keyframes by utilizing event-driven spiking mechanisms that capture fine-grained temporal dynamics. Furthermore, the F1-mean metric steadily increases, reaching its peak at three layers with a value of 19.9, suggesting that a deeper SNN architecture promotes more consistent keyframe selection across diverse video samples. However, the slight decline in 50.2 of F1-max at three layers implies that an excessive number of layers may introduce redundancy or reduce the model’s generalization ability.

For the TVSum dataset, F1-max scores remain relatively high even without SNN layers, indicating that the dataset’s inherent characteristics enable the model to perform well without extensive spiking neural computations. Interestingly, adding one or two layers slightly reduces performance, with F1-max decreasing to 77.9 and 78.6, respectively. This decline suggests that the dataset benefits more from conventional feature extraction than from the event-driven nature of SNNs. However, F1-mean peaks at two layers, reaching a score of 57.4, illustrating that a moderate SNN depth enhances the stability and balance of keyframe selection.

\begin{table}
	\centering
	\begin{threeparttable}
		\renewcommand\arraystretch{1.2}
		\caption{Ablation Study on Different Spiking Neuron.}
		\label{TAB6}
		\setlength{\tabcolsep}{5pt}{
\begin{tabular}{cccccccccccccccc}
\hline \hline
  \multirow{2}{*}{Neuron} &
  \multicolumn{3}{c}{SumMe} &&
  \multicolumn{3}{c}{TVSum} &
   \\ \cline{3-4} \cline{7-8} 
 &
   &
  F1-max $\uparrow$&
  F1-mean $\uparrow$&
   &
   &
  F1-max $\uparrow$&
  F1-mean $\uparrow$&
   \\ \hline
   LIF &&
  59.9 &
  17.8 &
   &
   &
   78.6 &
   57.4 &\\
   IF &&
  49.7 &
  21.8 &
   &
   &
  78.9 &
  56.3 &\\
  QIF &&
  47.0 &
  22.1 &
   &
   &
   80.5 &
   57.2 &\\ 
  EIF &&
  47.6 &
  19.9 &
   &
   &
   75.9 &
   56.0 &\\
  \hline \hline
\end{tabular}
}
\end{threeparttable}
\end{table}

\subsubsection{The Effectiveness of Different Spiking Neurons}
Table \ref{TAB6} presents an ablation study comparing the performance of various spiking neuron models on the SumMe and TVSum datasets. On the SumMe dataset, the Leaky Integrate-and-Fire (LIF) neuron achieves the highest F1-max score of 59.9, demonstrating strong keyframe detection capabilities for effective video summarization. However, its F1-mean score of 17.8 indicates variability in selection consistency. In contrast, the Integrate-and-Fire (IF) neuron records a lower F1-max of 49.7 but a notably higher F1-mean of 21.8, suggesting that while it may not always capture the most optimal keyframes, it ensures more consistent selection across different sequences. The Quadratic Integrate-and-Fire (QIF) and Exponential Integrate-and-Fire (EIF) neurons exhibit comparable F1-mean scores of 22.1 and 19.9, respectively, but lower F1-max values of 47.0 and 47.6. The QIF neuron, incorporating a quadratic voltage-dependent term, appears to enhance keyframe selection stability compared to the EIF neuron, which employs an exponential voltage term to mimic biological firing behavior.
 
For the TVSum dataset, performance trends shift, underscoring the dataset-specific influence of neuron models. The QIF neuron achieves the highest F1-max of 80.5, suggesting that its quadratic integration mechanism aligns well with the dataset’s temporal structure, particularly in handling gradual scene transitions. These findings highlight the significant impact of neuron type on video summarization performance, with LIF excelling in identifying peak keyframes in datasets like SumMe, while QIF proves more effective for datasets with smoother transitions like TVSum.

\begin{figure*}
	\centering
	\includegraphics[width=1\linewidth]{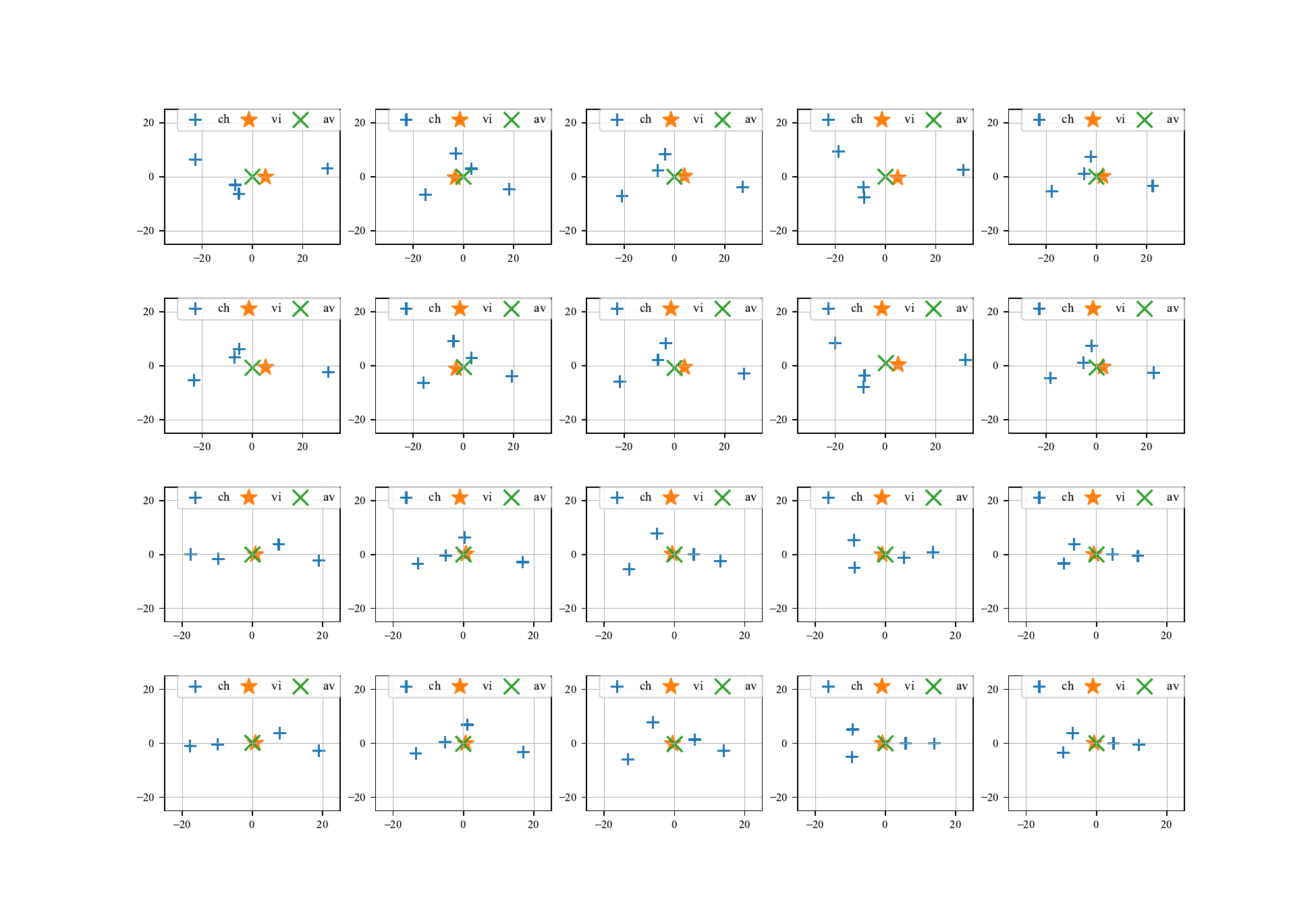}
	\caption{Visualization results of PCA and Isomap for partial videos on SumMe and TVsum datasets, including summary results (ch), mean values (av), and results obtained from the Variational Inference Reconstruction Module (vi).}
	\label{results}
\end{figure*}
\subsubsection{The Effectiveness of Different Components}
To further investigate the standalone contributions of key components in SpiVG, we conducted ablation experiments on the QFVS dataset by selectively removing the Text-oriented Multi-modal Embedding and the Dynamic Aggregation Graph Reasoner, respectively. As shown in Table \ref{ablation}, removing either component leads to a notable degradation in performance across all video categories. Specifically, the removal of the multi-modal embedding module results in a significant drop in average F1-score from 50.20 to 32.61, indicating that the textual guidance plays a crucial role in aligning semantic content with user intent. On the other hand, removing the graph reasoning module leads to a reduction in F1-score to 43.26, highlighting the module’s effectiveness in capturing inter-frame semantic coherence and contextual object consistency. These results demonstrate that both components contribute complementary strengths: while the multi-modal embedding enhances query relevance, the graph reasoning module ensures structural and semantic consistency. Together, they form a synergistic foundation for producing high-quality, user-adaptive video summaries.

\subsection{The Visualization Analysis of PCA and Isomap}
Fig. 4 provides a comprehensive visualization of the Variational Inference Reconstruction Module (VIRM) using two different dimensionality reduction techniques: Principal Component Analysis (PCA) and Isomap. These methods are applied to analyze the SumMe and TVSum datasets, enabling an in-depth comparison of the spatial distribution of different summary representations. Specifically, the figure maps the relationships among three key components: the original summary results (denoted as “ch”), the VIRM-generated summaries (denoted as “vi”), and the mean values computed across the dataset (denoted as “av”). By reducing the feature space into lower-dimensional embeddings, the visualization highlights how effectively VIRM reconstructs the patterns of the original summaries compared to a mean-based approach.

A crucial observation from the visualization is the consistent overlap between the “ch” and “vi” clusters across different embeddings. This suggests that VIRM accurately preserves the original summaries' key structures and semantic consistency, successfully capturing essential temporal and contextual information. In contrast, the “av” points appear more widely dispersed, indicating that averaging keyframe selections results in a loss of meaningful structure. The increased spread of “av” suggests that simple averaging fails to account for the nuanced dependencies between frames, leading to a less informative summarization. These findings underscore the effectiveness of VIRM in reconstructing video summaries while maintaining their underlying patterns, making it a superior alternative to conventional averaging techniques.
\begin{table*}
	\centering
	\begin{threeparttable}
		\renewcommand\arraystretch{1.4}
		\caption{Ablation experiment on QFVS Dataset.}
		\label{ablation}
		\setlength{\tabcolsep}{3pt}{
\begin{tabular}{ccccclccclccclccclccclccc}
\hline \hline
  &
  \multirow{2}{*}{Model} &
  \multicolumn{4}{c}{Vid 1} &
  \multicolumn{4}{c}{Vid 2} &
  \multicolumn{4}{c}{Vid 3} &
  \multicolumn{4}{c}{Vid 4} &
  \multicolumn{3}{c}{Average} &
   \\ \cline{3-5} \cline{7-9} \cline{11-13} \cline{15-17} \cline{19-21} 
 &
   &
  P $\uparrow$ &
  R $\uparrow$ &
  F1 $\uparrow$&
  &
  P $\uparrow$&
  R $\uparrow$&
  F1 $\uparrow$&
  &
  P $\uparrow$&
  R $\uparrow$&
  F1 $\uparrow$&
  &
  P $\uparrow$&
  R $\uparrow$&
  F1 $\uparrow$&
  &
  P $\uparrow$&
  R $\uparrow$&
  F1 $\uparrow$&
   \\ \hline
    & w/o Multi-modal Embedding
    & 49.41 
    & 28.34 
    & 36.02 &
    & 41.49 
    & 45.12 
    & 43.23 &
    & 50.10 
    & 24.92 
    & 33.28 &
    & 10.48 
    & 61.31 
    & 17.90 &
    & 37.87 
    & 39.92
    & 32.61 & \\
    & w/o Dynamic Graph Reasoner
    & 52.51 
    & 33.65 
    & 41.02 &
    & 42.14 
    & 49.24 
    & 45.41 &
    & 66.31 
    & 46.34 
    & 54.55 &
    & 22.44 
    & 56.14 
    & 32.06 &
    & 45.85
    & 46.34
    & 43.26 & \\\hline
    & SpiVG (ours)
    & \textbf{60.26}
    & \textbf{43.32}
    & \textbf{49.29} &
    & \textbf{49.59} 
    &  \textbf{56.64}
    & \textbf{52.48} &
    &\textbf{70.85}  
    & \textbf{51.93}
    & \textbf{59.62} &
    & \textbf{29.91}   
    & \textbf{59.66}    
    & \textbf{39.41} &
    & \textbf{52.65}  
    &  \textbf{52.89}
    & \textbf{50.20} & 
  \\
  \hline \hline
\end{tabular}
}
\end{threeparttable}
\end{table*}

\begin{table}
	\centering
	\begin{threeparttable}
		\renewcommand\arraystretch{1.2}
		\caption{The computational cost comparison in the TVSum dataset.}
		\label{TAB7}
		\setlength{\tabcolsep}{5pt}{
\begin{tabular}{c|ccc|c}
\hline \hline
Model   &$\tau$ $\uparrow$&
  $\rho$ $\uparrow$ & \#Param (MB) $\downarrow$ & Total (MB) $\downarrow$ \\ \hline
PGL-SUM \cite{apostolidis2021combining}    &0.27 &0.39   & 36.02   & 55.17            \\ 
A2Summ \cite{he2023align}     &0.26 &0.38 & 9.60    & 50.56           \\ 
VideoSAGE \cite{chaves2024videosage}  &\textbf{0.30} &0.42 & 3.52   & 19.27            \\ \hline
SpiVG(ours)  &\textbf{0.30} &\textbf{0.43} & \textbf{2.97}   & \textbf{18.58}            \\\hline \hline
\end{tabular}}
\end{threeparttable}
\end{table}

\begin{figure*}
	\centering
	\includegraphics[width=1\linewidth]{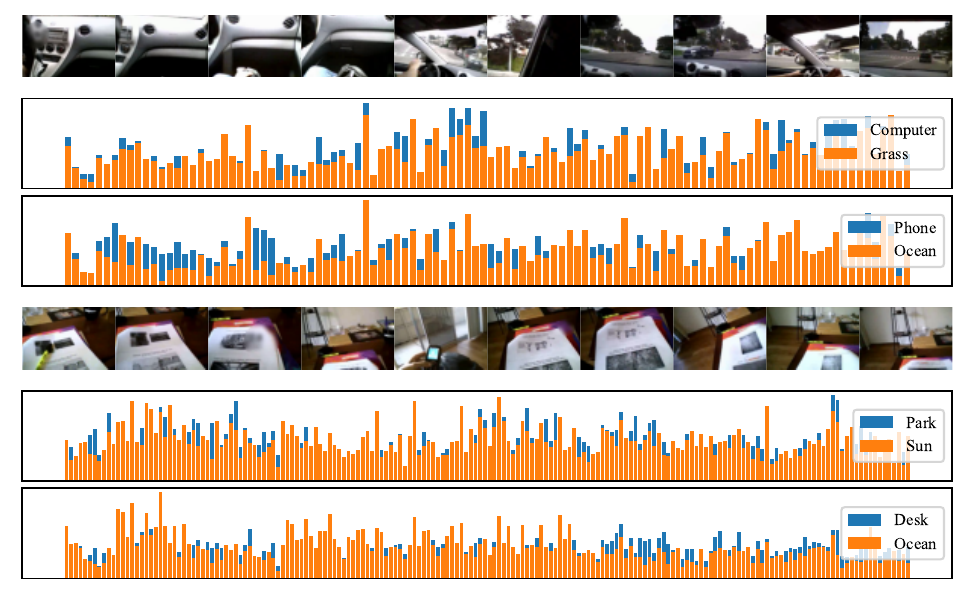}
	\caption{The summary score of each frame made by SpiVG under certain videos and given natural language instructions, which describes the likelihood of the frame becoming a summary frame. It can be seen that the summary results will change with the instructions, indicating that the model has effectively fused multiple modalities.}
	\label{results}
\end{figure*}

\subsection{Computational Efficiency}
Table \ref{TAB7} provides a comparative analysis of the computational efficiency of various models on the TVSum dataset, evaluating key factors such as ranking consistency, parameter size, and memory usage. The results highlight the efficiency of the Spiking Variational Graph (SpiVG) Network, which achieves an optimal balance between performance and resource consumption. Specifically, SpiVG achieves high-ranking consistency with a $\tau$ of 0.30 and a $\rho$ of 0.43, outperforming or matching other models.

Regarding computational efficiency, SpiVG significantly outperforms other models by maintaining the smallest parameter size of just 2.97 MB and the lowest memory usage at 18.58 MB. This compact architecture considerably reduces computational costs compared to larger models such as PGL-SUM and A2Summ, which require substantially more resources. Notably, SpiVG surpasses VideoSAGE, a model designed for efficiency, further underscoring its ability to deliver high-quality video summarization while minimizing computational overhead. These findings emphasize the advantage of SpiVG in achieving state-of-the-art summarization performance without the trade-off of excessive memory or parameter requirements, making it a more scalable and resource-efficient solution for real-world applications.

\subsection{Quantitative Analysis}
Fig. 5 presents a detailed analysis of SpiVG’s ability to retrieve relevant video segments based on different query pairs, demonstrating its effectiveness in query-focused video summarization. Each bar chart visualizes the relevance scores assigned to two competing query terms across video frames, illustrating how the model differentiates between semantic categories. The results highlight SpiVG’s capacity to dynamically adjust relevance scores based on contextual cues, ensuring accurate frame-query associations.

A more detailed examination reveals distinct patterns in SpiVG’s score assignments across varying scene contexts. For instance, the query term “Computer” consistently receives higher relevance scores in indoor settings, where objects like desks, monitors, and keyboards are present. At the same time  “Grass” dominates in outdoor scenes featuring natural landscapes. Similarly, the model assigns higher relevance to “Phone” in indoor frames, where handheld devices are more common, whereas “Ocean” emerges as the dominant term in maritime or beach-related scenes. This contrast in relevance scores across different settings underscores SpiVG’s ability to interpret video content accurately, capturing temporal variations and adapting its retrieval mechanism to real-world contextual cues.

\section{Conclusion}
\label{sec:conclu}
In this work, we present the Spiking Variational Graph (SpiVG) Network, a novel framework that integrates Spiking Neural Networks (SNNs) with variational graph reasoning for video summarization.
Our approach focuses on multi-channel feature fusion, maintaining contextual and semantic consistency, and achieving robust summarization.
The Dynamic Aggregation Graph Reasoner separates contextual object consistency from semantic viewpoint coherence, enabling fine-grained inter-frame reasoning. 
The Variational Inference Reconstruction Module minimizes uncertainty during feature integration, ensuring that the final summaries are semantically rich and computationally efficient.
Our model consistently outperforms state-of-the-art methods across multiple datasets, including SumMe, TVSum, and QFVS.


\bibliographystyle{unsrt}

\end{document}